\documentclass[sponsored]{acmsiggraph}

\usepackage{subfig}
\usepackage{url}
\usepackage{gensymb}

\title{Rendering refraction and reflection of eyeglasses for synthetic eye tracker images}

\author{Thomas C. K\"ubler\thanks{e-mail: thomas.kuebler@uni-tuebingen.de}\\ Tobias Rittig\\ Enkelejda Kasneci\\ Wilhelm-Schickard-Institute of Computer Science\\ University of T\"ubingen%
\and Judith Ungewiss\thanks{e-mail:judith.ungewiss@hs-aalen.de}\\ Christina Krauss\\  Study Course Ophthalmic Optics/Audiology\\
University of Applied Sciences Aalen}

\pdfauthor{Thomas C. K\"ubler, Tobias Rittig, Judith Ungewiss, Christina Krauss, Enkelejda Kasneci}


\begin{document}




\maketitle


\begin{abstract}
While for the evaluation of robustness of eye tracking algorithms the use of real-world data is essential, there are many applications where simulated, synthetic eye images are of advantage. They can generate labelled ground-truth data for appearance based gaze estimation algorithms or enable the development of model based gaze estimation techniques by showing the influence on gaze estimation error of different model factors that can then be simplified or extended.
We extend the generation of synthetic eye images by a simulation of refraction and reflection for eyeglasses.
On the one hand this allows for the testing of pupil and glint detection algorithms under different illumination and reflection conditions, on the other hand the error of gaze estimation routines can be estimated in conjunction with different eyeglasses.
We show how a polynomial function fitting calibration performs equally well with and without eyeglasses, and how a geometrical eye model behaves when exposed to glasses.
\end{abstract}


\begin{CRcatlist}
  \CRcat{I.6.5}{Simulation and Modeling}{Model Development—Modeling methodologies};
	\CRcat{I.3.8}{Computer Graphics}{Applications};
\end{CRcatlist}


\keywordlist




\copyrightspace


\section{Introduction}

Video-oculography is based on two computational steps: detecting features such as the pupil or glints on the cornea surface, and gaze mapping, i.e., the transformation of image features into a gaze direction. For the latter step, one can distinguish between function fitting, model-based and appearance-based algorithms~\cite{villanueva2007models}. Appearance-based methods learn what certain gaze directions look like in the image from a huge amount of training samples, i.e., a large amount of images with annotated gaze direction. In order to efficiently generate such training data there have been approaches to generate synthesized images as a labeled ground-truth~\cite{wood2015rendering,sugano2014learning}.
Alternatively, a mapping function can be used in order to transform image features into a gaze direction. A polynomial of order two is a common choice in order to map the mid of the pupil in the eye image to gaze coordinates in a calibration plane. Therefore the subject is required to look at some locations in the calibration plane. The point correspondences between these known locations and the respective pupil center coordinate in the eye image are then used to adjust the parameters of the mapping function.

\begin{figure}[ht]
  \centering
  \includegraphics[width=1.5in]{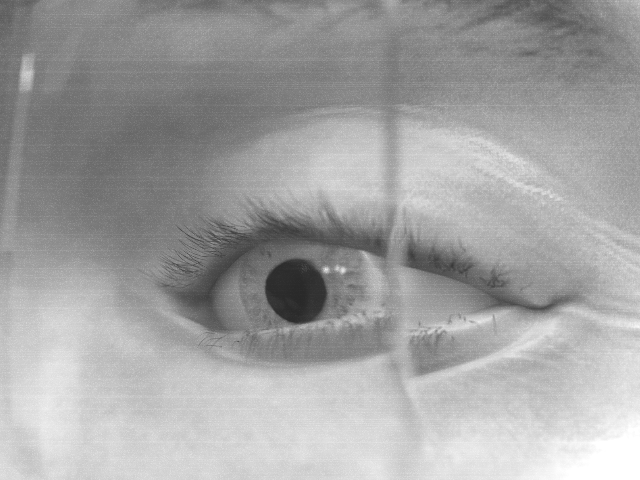}
  \caption{Rendered image of an eye with a coated lens of -1 diopters.}
\end{figure}

Even though the above approaches are common and can reach high accuracy, there are applications where such time-consuming and error-prone calibration processes are not applicable, for example when tracking the eye movements of children or patients with impaired ocular motility. Therefore, there is continuous effort to employ model-based gaze mapping techniques. Such models usually include representations of the human eye, IR light sources and the camera. The assumption of a model reduces the amount of parameters that require adjustment through the calibration process and population averages can be used or adjusted over time in order to work completely calibration free.
While most model-based methods employ at least two IR LEDs and often multiple cameras (e.g., using the shape of the pupil and the center of the cornea~\cite{villanueva2007models,hnatow2006efficient}), there are also approaches with one camera and no IR reflections~\cite{swirskifully}.

To the best of our knowledge, there is no geometrical model  including a representation of eyeglasses. Consequently, gaze mapping techniques are usually evaluated on healthy-sighted subjects. However, considering that about 30 percent of young adults in industrial nations need to wear eyeglasses~\cite{morgan2005genetic,schaeffel2006myopia}, evaluation techniques need to include such subject in their study populations. Another subject population that is rarely tracked in eye-tracking studies are elderly subjects. Indeed, this group  is often underrepresented in university studies, where young students are easy to recruit (with students having four times higher risk for myopia than persons with only primary schooling~\cite{morgan2005genetic}). In the group of elderly subjects, however, the prevalence of eyeglasses, especially of those with more complicated varifocals design, is even higher than in the rest of the population.

\begin{figure}[ht]
  \centering
  \includegraphics[width=2in]{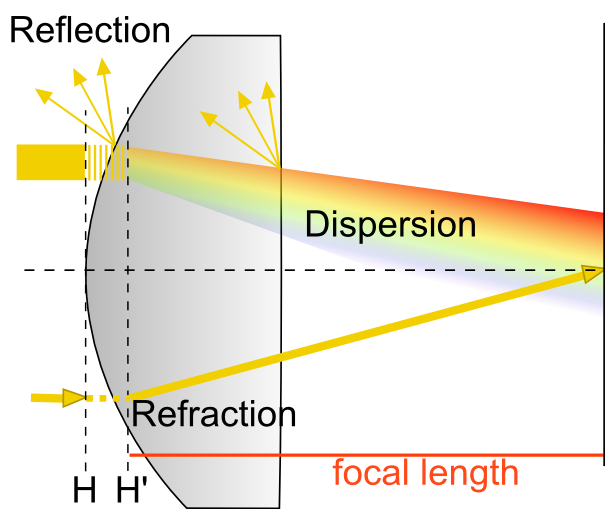}
  \caption{Most relevant effects of an eyeglass are refraction and reflection. Refraction is the effect of direction change of the light ray determined by the refractive index of the lens material as well as the thickness of the lens. Different wavelengths of light undergo refraction of different strength, an effect called dispersion. Reflective properties depend on the material as well as on the glass coating and its effectiveness for a certain wavelength of the light.}
	\label{glasses}
\end{figure}

Intuitively, the introduction of eyeglasses should not have a major impact on the accuracy of the function fitting technique for gaze estimation. Pupil as well as gaze positions are measured through the glasses during both calibration and measurement. Therefore, the parameters of the fitted function are adjusted to incorporate the effect of the eyeglasses. Geometrical models, however, are based on calculating a ray from the pupil center towards the camera. This ray is refracted by the eyeglasses (Figure~\ref{glasses}), introducing an error that is not accounted for by the model and that depends on the strength of the optical medium. This will result in a change of pupil center location as well as of the elliptical shape of the pupil.

\begin{figure}[ht]
  \centering
  \includegraphics[width=1.5in]{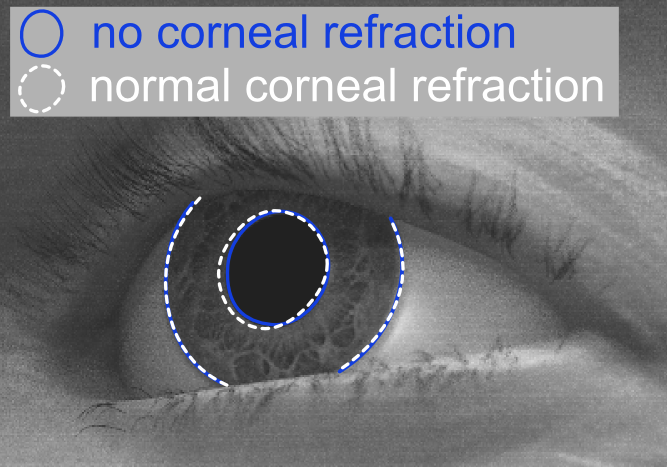}
  \caption{Effect of the corneal refraction on the pupil contour. Simulated image with normal refractive index of the cornea (1.336) and with refractive index equal to the air, causing no refraction.}
	\label{cornea}
\end{figure}

Significant inaccuracies in gaze ray tracing due to refraction can also be observed in relation with the cornea. Light rays towards the pupil are refracted by the cornea, causing the image of the pupil to appear different to the camera than without the refraction, see Figure~\ref{cornea}. A gaze  mapping algorithm that compensates for this effect was introduced in~\cite{villanueva2007models}. More specifically, \cite{villanueva2007models} quantify the gaze error when ignoring corneal refraction at about 3~\degree.
The appearance change of the pupil caused by refraction of the cornea can be avoided by detecting and tracking the iris contour instead of the pupil as it is done in~\cite{wang2005estimating}. As can be seen in Figure~\ref{cornea} the effect of corneal refraction is much smaller for the iris. However, the iris is harder to be automatically detected and tracked. However, this stability of the iris contour does not hold for eyeglasses.

There are purely geometrical eye models that were specifically developed in order to provide ground-truth data sets for eye tracker development and evaluation. For example, \cite{bohme2008software} developed a geometrical model that could determine the positions and orientations of camera, eye, and pupil.

Another important step to enable robust and easy eye-tracking in real-world settings is handling variations in illumination. While this is a challenge solely because of the anatomical shape and dynamics of the eye region, eyeglasses enforce this challenge due to additional reflections. While laboratory conditions allow for the detection of a large number of IR reflections on the cornea, real-world environments with uncontrolled, natural illumination often make finding even one reflection quite a challenging task. The reflective surface of the eyeglasses will produce an image of the IR illumination, just like the cornea does.

Our work is based on the work of {\'S}wirski and Dodgson~\cite{swirski2014rendering}. The authors render a 3D model of the eye region, and extend the model of~\cite{bohme2008software} by the image generation step. Our approach utilizes the model and rendering pipeline provided by~\cite{swirski2014rendering} and produces highly realistic images of the eye with different eyeglasses as it would appear on the image of a mobile eye tracker with IR illumination.
Thereby we provide ground truth data to evaluate the influence of different lenses on gaze mapping accuracy.

\section{Model}
\subsection{Modeling of the lens}
Our model provides code to generate the 3D shape of a simple plano-concave/convex lens with arbitrary diopter. The lens shape is cut from a box-sphere intersection resulting in one curved and one planar surface. Positive diopters require a converging lens, modeled as a plano-convex lens with the planar surface towards the eye. Plano-concave lenses (negative diopters) are modeled with the curved surface towards the eye.

The refractive index of the lens material can be adjusted and we provide meaningful defaults for materials currently used in the manufacturing of eyeglasses (such as N-BK7 Schott). Since most head-mounted eye trackers work with near infrared illumination, we chose the refractive index at near infrared (900~nm) for rendering. During calculation of the curvature radii of the lens the refractive index at 589~nm (Fraunhofer D line) is taken into account.

Modeling the reflective properties of a lens via ray tracing is complex for reflex reducing coated lenses. Therefore, the reflection process for these lenses is only an approximate based on measurements of the overall reflectiveness of such lenses~\cite{zeiss2002wissenswertes}. It is notable that the efficiency of reflex reduction is wavelength dependent and not optimized for infrared conditions. Therefore, the eye tracker is likely to encounter more reflections than what humans will be able to see.
In contrast to {\'S}wirski and Dodgson's model, correct reflections on the eyeglass surfaces require an environment to reflect. Therefore, we added an environment map as in~\cite{hdrenvmap} that is adjusted online in order to resemble the infrared illumination. Limited by the availability of infrared HDR environment maps, we chose to simulate by a pseudo-infrared tone mapping technique taken from a preset of Adobe Photoshop.

\begin{figure}
\begin{tabular}{ccc}
\subfloat{\includegraphics[width = 1in]{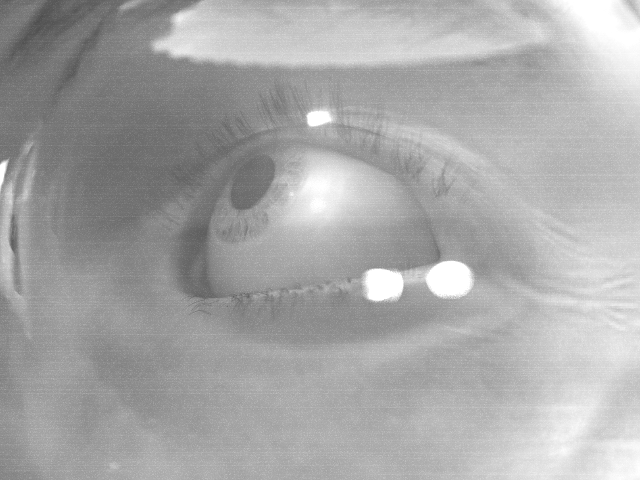}} &
\subfloat{\includegraphics[width = 1in]{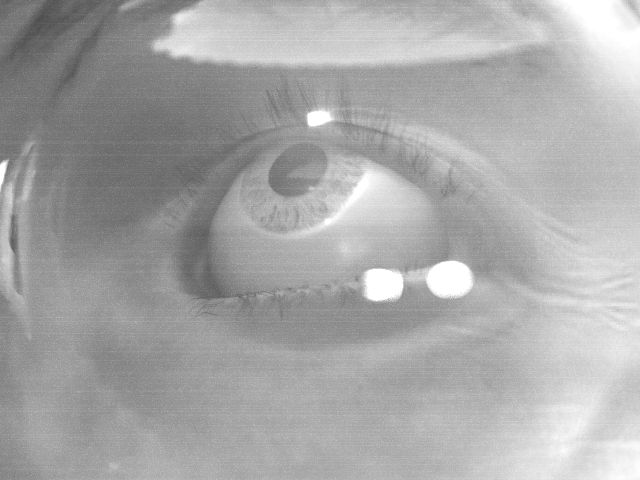}} &
\subfloat{\includegraphics[width = 1in]{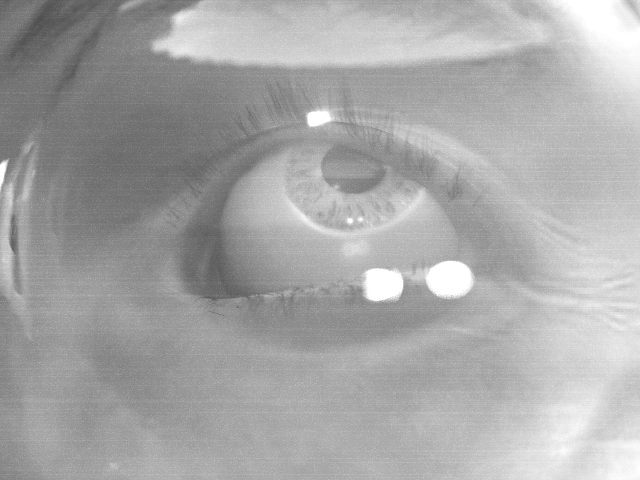}} \\
\subfloat{\includegraphics[width = 1in]{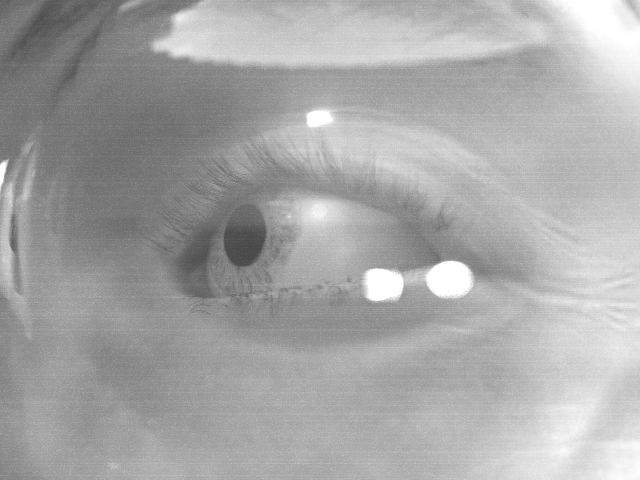}} &
\subfloat{\includegraphics[width = 1in]{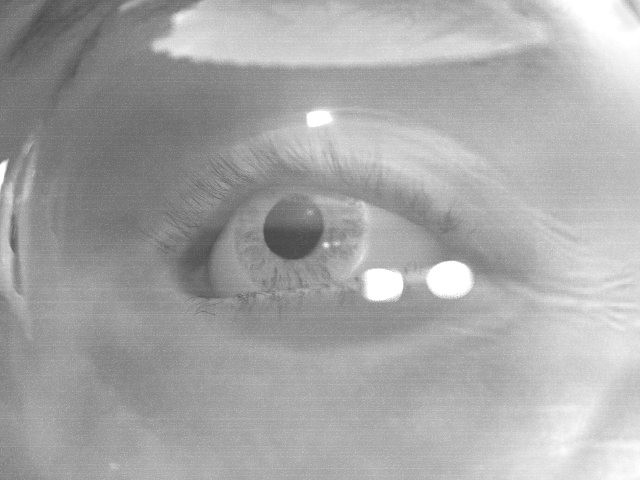}} &
\subfloat{\includegraphics[width = 1in]{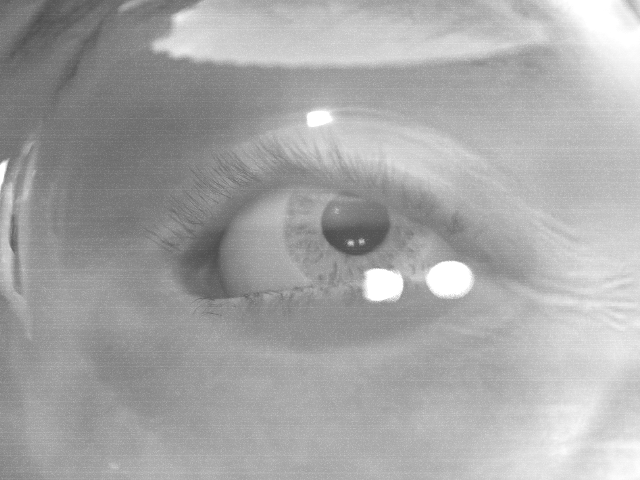}} \\
\subfloat{\includegraphics[width = 1in]{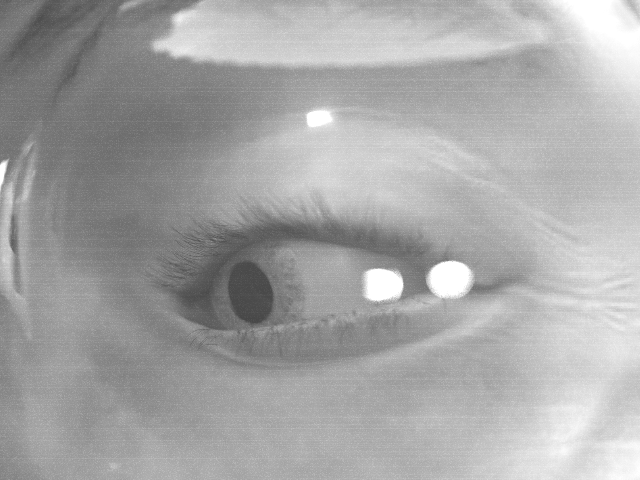}} &
\subfloat{\includegraphics[width = 1in]{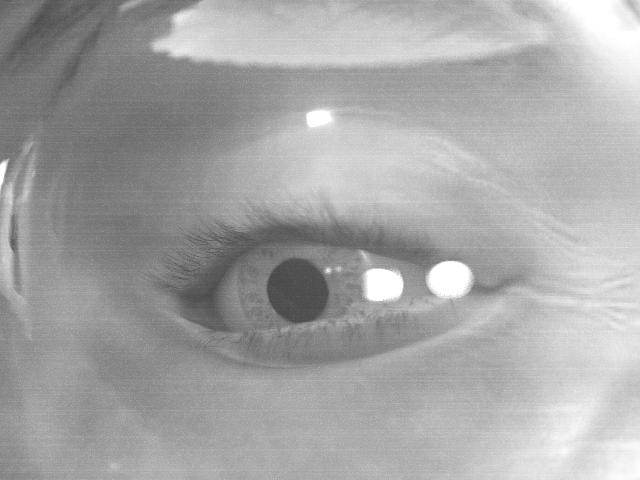}} &
\subfloat{\includegraphics[width = 1in]{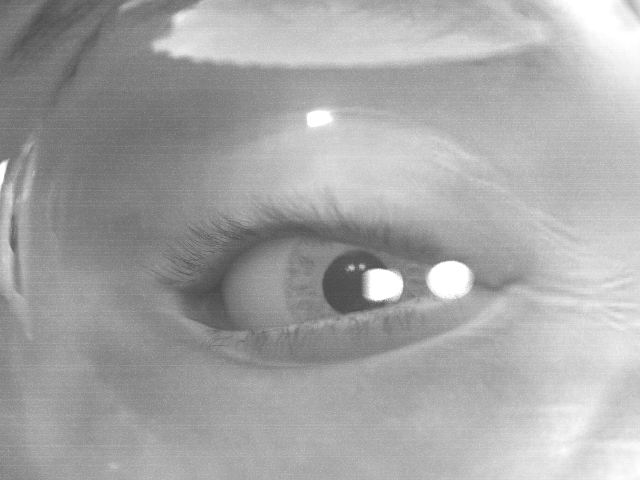}} \\
\end{tabular}
\caption{Generated calibration images with a -1 dpt uncoated lens. Gaze directions vary between -20 and 20~\degree horizontally and vertically.}
\label{calibrationimages}
\end{figure}

\subsection{Gaze mapping accuracy}
In order to determine the influence of glasses on gaze mapping algorithms we generated images for a calibration step. For the polynomial calibration, we used nine images (Figure~\ref{calibrationimages}) in primary (straight-ahead), secondary (up-down/left-right) and tertiary (diagonal) position at 20\degree. The calibration accuracy based on the polynomial fit is usually highest close to the calibration points and decreases with distance. This is especially true for synthetic images. Therefore, we generated an additional test image set with 16 more locations distributed within the calibrated area. In the recorded images, we annotated the pupil boundaries by fitting an ellipse to ten points on the pupil edge. These point were manually annotated. Thus, we ensure that all inaccuracies are caused by the gaze prediction step, not by an insufficiently annotated pupil.

For a geometrical eye model, there is theoretically no need for a calibration. However, we used five points in the primary and secondary position in order to determine the gaze ray in relation to the x- and y-axis defined by the secondary positions. This allows for a correction of the translation effects caused by the eyeglasses. The original code of the geometrical model was used where possible as presented in~\cite{swirskifully}. The annotation of the pupils, ellipse fitting, as well as post-processing by selection of the x- and y-axis were programmed around the original library. It should be noted that we took this model as an example, however there are several other geometrical eye models that could be employed. None of them models eyeglasses and many require additional glint points on the cornea that cannot be determined as easily when the subject is wearing eyeglasses as without. Furthermore, we expect a similar effect on gaze prediction accuracy for other models.

\begin{figure}[ht]
  \centering
  \includegraphics[width=3.3in]{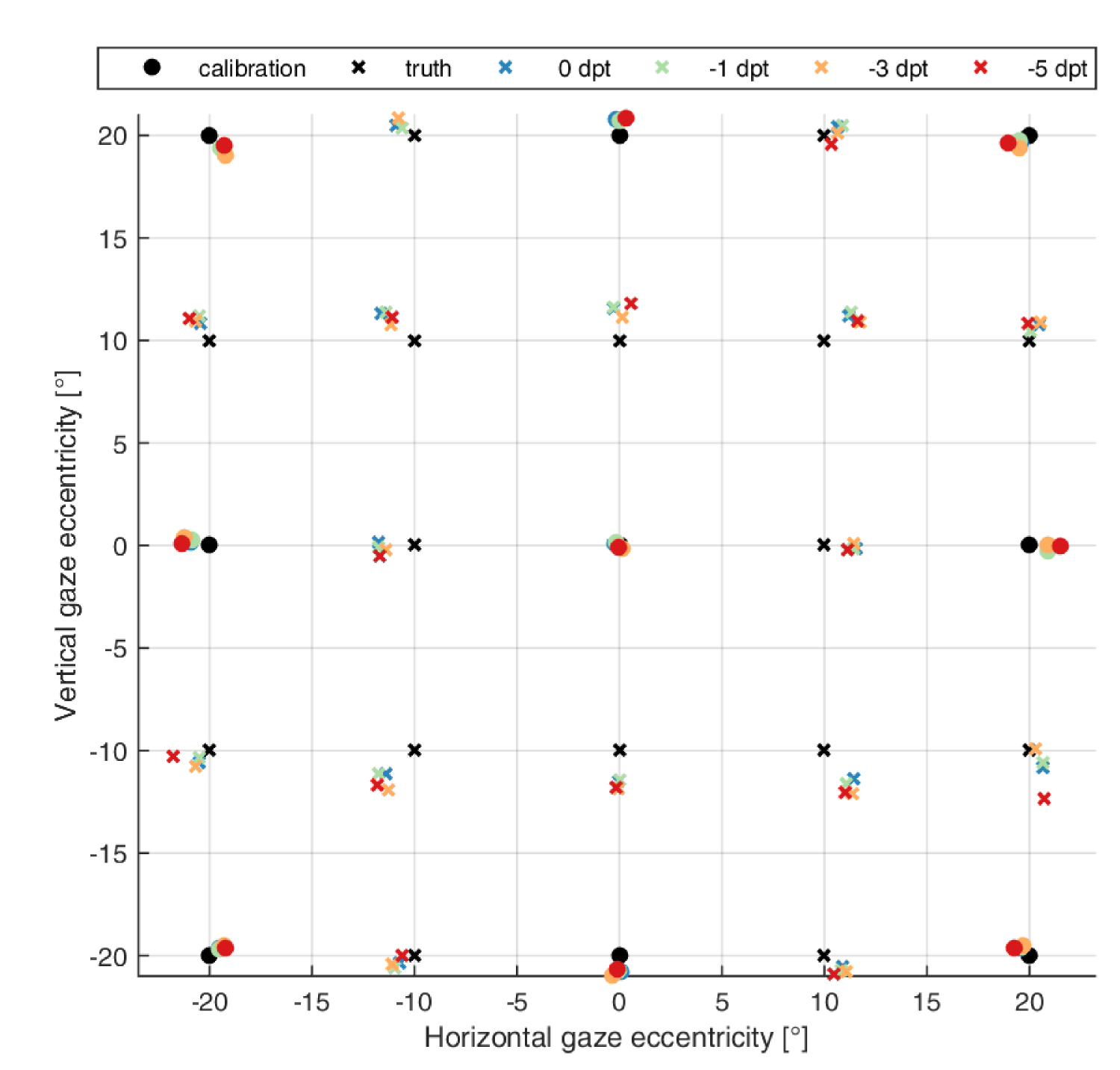}
  \caption{Polynomial gaze mapping for different eyeglasses (up to -5 dpt) and without glasses (0 dpt). Points at locations marked by black circles were used for calibration. Black crosses denote the true gaze orientation at the test points. No significant effect of eyeglasses on the calibration accuracy can be observed.}
	\label{polynomial}
\end{figure}

\begin{figure}[ht]
  \centering
  \includegraphics[width=3.3in]{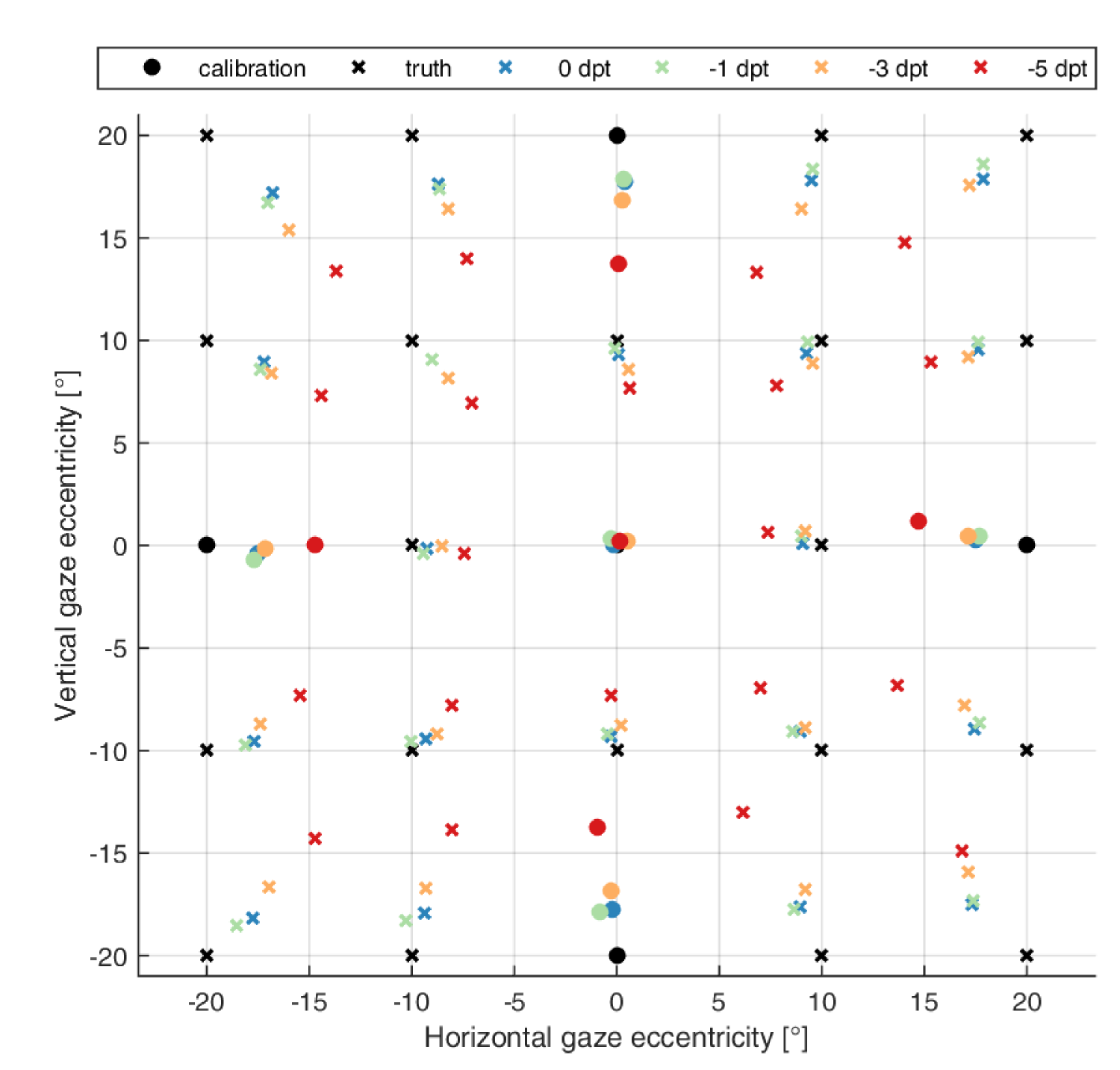}
  \caption{Geometrical gaze mapping for different eyeglasses (up to -5 dpt) and without glasses (0 dpt). Points at locations marked by black circles were used for calibration, whereas black crosses denote the true gaze orientation at the test points. A relevant decrease of gaze prediction accuracy coupled to the refractive strength of the eyeglasses can be observed.}
	\label{geomodel}
\end{figure}

Gaze mapping accuracy was measured as the angular distance between the actual gaze target and the predicted one. This procedure was applied to images of the eye without glasses, with weak lenses (-1 dpt), as well as stronger lenses (-3 dpt and -5 dpt). Figure~\ref{polynomial} shows the accuracy of the polynomial mapping, Figure~\ref{geomodel} the geometrical model. Numerical averages over all test positions are reported in Table~\ref{resulttable}.

\begin{table}
	\centering
		\begin{tabular}{ r c c }
			Eyeglass & Polynomial fitting & Geometrical model \\ \hline
			\hspace{1pt} 0 dpt & 1.33 $\pm$ 0.46~\degree & 2.09 $\pm$ 1.05~\degree \\ 
			-1 dpt & 1.33 $\pm$ 0.52~\degree & 1.95 $\pm$ 1.10~\degree \\ 
			-3 dpt & 1.37 $\pm$ 0.57~\degree & 2.94 $\pm$ 1.43~\degree \\ 
			-5 dpt & 1.56 $\pm$ 0.61~\degree & 5.38 $\pm$ 2.16~\degree \\  \hline
		\end{tabular}
	\caption{Mean accuracy ($\pm$ standard deviation) of gaze prediction for both calibration algorithms and different eyeglasses. While the polynomial fitting shows only a minor decease in average accuracy and a minor increase in standard deviation, the geometrical model is heavily influenced by the refractive strength of the eyeglasses.}
	\label{resulttable}
\end{table}

\section{Discussion}

We proposed a method for close to realistic rendering of eyeglasses in synthetic eye-tracking images. Effects of refraction and reflection can be studied in connection with pupil and glint detection as well as gaze mapping techniques. We further showed that model-based gaze mapping techniques are sensitive to the refraction of eyeglasses.

Our model can already generate many different lens designs. However, there are certain limitations to this work with regard to the design of the lenses. Up to now, we have focused on planar-concave/convex lenses. An extension for convex-concave lenses as well as additional optical effects such as prisms and cylinder is under development. Therefore, in this work, we only use a small subset of all possible glass designs. Creating a representative catalog of frequently used eyeglasses is a desirable next step of our future work.

Due to the more complex calculations and increased number of ray samples required for the refraction and reflection effects, rendering time has increased significantly when compared to the model without glasses. Blender's cycles renderer implements a path tracing algorithm and thus is unsuitable for these kind of complex reflections. Multiple importance sampling does not improve convergence in this scenario due to the refraction of each ray in the lens separating the skin surface from the light sources. Our planned improvement in rendering time is achieved using a different rendering engine such as Luxrender, which implements bidirectional pathtracing. This rendering algorithm connects both camera and light paths at each bounce and thus converges with less samples.

The approximation of the environment map in infrared light is in our opinion not critical due to the lens coating reflecting only \~5\% of the irradiance. Specular highlights of the environment are thus only clearly visible in bright areas such as the sky. Comparing near infrared and visible light brightness in solar radiation does reveal significant differences due to atmospheric absorption\cite{lacis1974parameterization} but they are unlikely to be noticeable within the dynamic range of an eye-tracker camera.

There are several possible applications, where eye trackers could be used over long periods of time. An interesting application context regards the autonomous driving, were tracking the eye movements of the driver is not only a good means to determine the driver's level of attention and estimate her capability to take control over the vehicle~\cite{braunagel2015exploiting}, but also in the context of human-machine-interaction. For such scenarios, we plan to extend the rendering model by a realistic simulation of dust and dirt, both on the eye-tracker's camera lens, as well as on the eyeglasses. This will have an effect on detection rate and accuracy.



\bibliographystyle{acmsiggraph}
\bibliography{eyeglasses}
\end{document}